# An Improved Deep Belief Network Model for Road Safety Analyses

Guangyuan Pan[1], Liping Fu*[12], Lalita Thakali[1], Matthew Muresan[1], Ming Yu[3]

[1]Department of Civil & Environmental Engineering,
University of Waterloo,
Waterloo, ON, Canada, N2L 3G1;
lfu@uwaterloo.ca

[2]Intelligent Transportation Systems Research Center,
Wuhan University of Technology,
Mailbox 125, No. 1040 Heping Road, Wuhan, Hubei, China. 430063;

[3]Department of Electrical & Computer Engineering,
University of Waterloo,
Waterloo, ON, Canada, N2L 3G1.

*Abstract*—Crash prediction is a critical component of road safety analyses.  A widely adopted approach to crash prediction is application of regression based techniques.  The underlying calibration process is often time-consuming, requiring significant domain knowledge and expertise and cannot be easily automated. This paper introduces a new machine learning (ML) based approach as an alternative to the traditional techniques.  The proposed ML model is called regularized deep belief network, which is a deep neural network with two training steps: it is first trained using an unsupervised learning algorithm and then fine-tuned by initializing a Bayesian neural network with the trained weights from the first step.  The resulting model is expected to have improved prediction power and reduced need for the time-consuming human intervention.  In this paper, we attempt to demonstrate the potential of this new model for crash prediction through two case studies including a collision data set from 800 km stretch of Highway 401 and other highways in Ontario, Canada.  Our intention is to show the performance of this ML approach in comparison to various traditional models including negative binomial (NB) model, kernel regression (KR), and Bayesian neural network (Bayesian NN).  We also attempt to address other related issues such as effect of training data size and training parameters.

## I. Introduction

According to a recent report by World Health Organization (WHO) on road safety, road traffic collisions account for approximately 1.25 million human fatalities and over 50 million non-fatal injuries every year worldwide [1]. The consequences of these effects are felt in every area of society, and cause an estimated 1% loss of gross national product (GNP) for low-income counties, 1.5% for middle-income countries and 2% for high-income countries [2]. Road safety has therefore become a top priority for many jurisdictions around the world, and significant investments have recently been made into various efforts aimed at understanding and improving road safety.

A critical step in road safety studies, whether it be in the identification of crash hotspots or in studying the effectiveness safety countermeasures, is the need to develop collision prediction models. Several approaches have been used in the past to develop collision models. Among them, the most popular approach is the model-based, also known as parametric approach, in which the crash occurrence is assumed to follow a specific probability distribution (e.g., Poisson, Negative Binomial) with its mean being estimated as a function of a set of predicting factors [3]. Although these parametric models are easy to understand and apply, the predicted results are often inaccurate due to the random nature of collision occurrences and the strong assumptions imposed by the distribution used to capture the randomness.

To address the limitation of the parametric models, some researchers have recently been exploring the use of machine-learning models, such as artificial neural networks (ANN), support vector machines (SVM) and multivariate adaptive regression spline (MARS), which have shown evidence of improved performance in terms of collision prediction [4-5]. However, these methods have some obvious disadvantages. Traditional neural networks are based on back propagation, which is a supervised training method that requires input from a teacher during the training process. Furthermore, it is very sensitive to the initial weights used, and poor initial values can trap the weights in a local minimum, resulting in sub-optimal values. Similarly, SVM models often use a kernel function to map input data into higher dimensional space for making predictions; however, the structure of SVM has been found to be limited in its ability to capture the effect of a large set of interrelated factors.

This paper expands on our previous work [6] where we showed that kernel regression, a non-parametric model, could be successfully employed to predict expected collision frequencies. The model was developed to take advantage of big data and to reduce testing errors [6]. In this paper, we propose a deep learning based model as an alternative technique to modeling collisions. Deep learning is one of the most recent and exciting techniques developed in machine learning [7]. It overcomes the shortcomings of traditional ANN and has been implemented successfully in many areas such as in pattern recognition, computer vision, and intelligent strategy [8-10]. In this paper, our focus is on solving the problem of predicting collisions for highways and achieving high predicting accuracy and fast training speed. We also explore the relation between the model size, data size and trained model accuracy.

This paper has the following structure: Chapter 2 reviews some of the popular collision prediction methods. Chapter 3 describes the model that we propose, and discusses the important parameters in training process. Chapter 4 presents two case studies using improved deep belief network. Finally, we conclude the study in chapter 5.

## II. LITERATURE REVIEW

The most extensively used approach for collision prediction in the past decades is generalized linear regression, including models such as the Poisson model, NB model, negative binomial model, Poisson Weibull model, Poisson-lognormal model, and zero-inflated negative binomial models [11-16]. These models are fundamentally alike and only differ in the distribution they assume for the collision occurrence. Because of the strong distribution assumption, these models may introduce significant bias or prediction errors under certain conditions [17-18]. Furthermore, these models are static in nature and time consuming to calibrate.

Within the field of computer science, machine learning has become a very popular topic of study in recent decades. Machine learning algorithms have also found their way into other fields, including road safety studies where they have been used for predicting collisions. For example, a recent study by Chang et al [19] implemented an artificial neural network (ANN) to predict collisions in National Freeway 1 in Taiwan. Machine learning techniques are computer-based techniques that aim to reproduce and simulate human behavior and cognitive functions. A number of different machine learning techniques have been proposed in the literature with ANN being one of the most popular ones. ANN uses a network of nodes (often called "neurons") containing configurable weights that can be trained to produce a desired output. These weights and layers can be configured to solve many kinds of pattern-recognition problems. For instance, in their study, Chang et al [19] set up an ANN model that accepts road condition features as input and provides collision numbers as output [20]. However, the solution space for ANN's weights is non-convex there are many local minimums that can trap the model as it is calibrated, making it difficult to find the global optimum solution. Another drawback of traditional ANN is its supervised learning construct that necessitates the availability of training data, limiting its applicability in real-world situations [21]. To address these problems, other improved versions of BP such as Bayesian regularization have been proposed by some researchers [22]. While Bayesian regularization shows improvements, it still requires training that limits its applicability. Other forms of statistical models used in collision predictions are kernel regression [6], hierarchical tree-based regression [23], multivariate adaptive regression splines (MARS), and support vector machines (SVM) [24-26]. They show promising results when road features and available data are limited. However, erecent technological advances have created a "big data" environment where data is abundant and easily collectable. These data sources capture significant details on individual collisions and their occurrence environment that are often not fully utilized by traditional models. Therefore, new models with greater prediction accuracy and faster speed are needed.

## III. REGULARIZED DEEP BELIEF NETWORK MODEL

Deep learning (DL) or deep neural network (DNN) is a new machine learning technique that has been widely explored and successfully applied for a variety of problems such as in image and voice recognition and games [27-28]. A few variations of DNN have been developed, among which the Deep Belief Network (DBN) is one of the most popular [9]. What makes DBN different is its unique training method called greedy unsupervised training. Its development is based on mimicking the cognitive and knowledge reference processes of the human brain. By stacking several Restricted Boltzmann Machines (RBM), which are a kind of recursive ANN model that contain two layers (one input layer and one output layer), DBN learns the features of input signals without needing the extra labelled data required in back propagation. Its configuration also allows it to obtain a better distributed representation of the input data. In this research, we explore the potential of applying DBN for road safety analysis, especially, prediction of collisions on highways.

Traditional Deep Belief Networks (DBN) can only process binary signals (0 and 1), however, real world road condition data and collision data are continuous (any value between 0 and 1). This limitation can be addressed by using an improved continuous version of the transfer function, as proposed by Bengio [32]. Contrastive Divergence (CD), a fast training method to train a DBN proposed by Hinton [33], is implemented for the unsupervised training process. The output of the first Restricted Boltzmann Machine (RBM) is set to be the input of the second RBM. The proposed model in this research also uses Bayesian regularization instead of traditional back propagation which raises prediction accuracy by reducing overfitting [34]. The proposed model is therefore called "Regularized DBN".

### A. Model Structure

An illustration of Regularized DBN model is shown in Figure 1. It consists of four layers. The first layer is the input layer (V) which receives the original feature data. This input layer is followed by two hidden layers, L1 and L2, with V and L1 forming the first RBM, and L1 and L2 the second RBM. The structure of each RBM is that of a two-way full connectivity between the two layers (as shown by double-sided arrows). No connections exist between units of the same layer. It is this absence of intra-layer connections that results in a so-called "restricted" Boltzmann Machine (RBM). In the training process, each hidden layer extracts the last layer's data information and features to form a better, although more abstract, distributed representation of input data. The last layer is the output layer which has only one unit, representing the predicted collision frequency. A structure with two layers was chosen since traditional DBN models employed in pattern recognition tasks often only include two layers. This structure strikes a balance between the need for a more powerful model and the desire to keep computation times lower. It offers more predictive power than a single layer model, but is significantly less time-consuming than a three layer model [27].

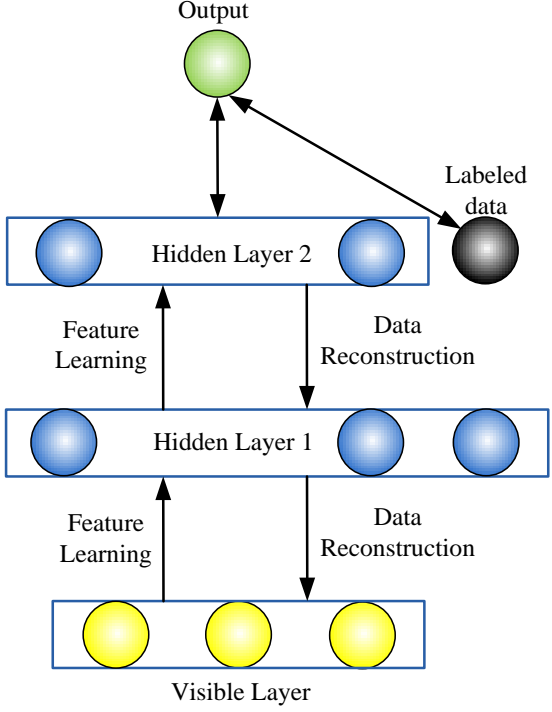

Fig.1. Regularized DBN's Structure

As a traditional DBN must transform the original signal to Bernoulli values (0 or 1) when processing the inputs, its applications are limited. Hinton [32] improved the input layer to receive continuous values, while the signal in the unsupervised training process will still be transformed to a Bernoulli value. According to previous work that has been done by Bengio [27], RBM is capable of processing continuous data and modeling in theory. Furthermore, some recent works have shown that continuous RBM can be used in continuous value prediction problem. For example, a unit-amount-selected continuous DBN is used in predicting a time series benchmark called CATS, good results are obtained [35]. Based on the works that have been done, the continuous version transfer function is employed.

*B. Model Algorithm*

Regularized DBN is made of several continuous RBMs, which are trained separately before unfolding all the layers and using Bayesian regularization to fine-tune the weights. Therefore, Regularized DBN combines unsupervised and supervised learning. The learning and reasoning processes in an RBM (also called knowledge generation) are shown in Equations (1) and (2).

$$p(h_j=1)=\frac{1}{1+e^{-b_j-\sum_i v_i w_{ij}}} \qquad (1)$$

$$p(v_i=1)=\frac{1}{1+e^{-c_i-\sum_j h_j w_{ji}}} \qquad (2)$$

where $v_i$ is the value of unit i of the input layer, $h_j$ is the value of unit j in the hidden (output of the RBM) layer; b and c stand for the biases of the input and hidden layers respectively; $w_{ij}$ is the weight between input unit i and hidden unit j. We then define $\theta=(W,b,c)$ according to a fast training method that commonly used, called Contrastive Divergence, to estimate successive new weights and biases with a Markov chain, as follows.

$$\theta^{(\tau+1)}=<h_j^0 v_i^0>-<h_j^1 v_i^1> \qquad (3)$$

where $<>$ denotes the average over the sampled states, $h_j^0 v_i^0$ is the initial state of the input layer multiplied by the hidden layer, and $h_j^1 v_i^1$ is the same product after a single iteration of a Markov chain.

In the proposed model, the transfer functions have been modified to process continuous data. Let $s_j$ and $s_i$ be the values of hidden unit j and input unit i, and the sigmoid function in Equations (1) and (2) will be kept, while the step of discretization is omitted. A noise function is also added to raise the ability of de-noising, and the final functions have the following form:

$$s_j=\varphi_j(\sum_i w_{ij}s_i+\sigma \cdot N_j(0,1)) \qquad (4)$$

$$s_i=\varphi_i(\sum_j w_{ij}s_j+\sigma \cdot N_i(0,1)) \qquad (5)$$

where,

$$\varphi_j(x_j)=\theta_L+(\theta_H-\theta_L)\cdot \frac{1}{1+e^{-a_j x_j}} \qquad (6)$$

$$\varphi_i(x_i)=\theta_L+(\theta_H-\theta_L)\cdot \frac{1}{1+e^{-a_i x_i}} \qquad (7)$$

Equations (4) and (5) are the processes of learning and generating, in which, N(0,1) is a Gaussian random variable with mean 0 and variance 1. $\sigma$ is a constant, and $\varphi()$ denotes the sigmoid function with asymptotes $\theta_H$ and $\theta_L$. a is a variable that controls noise, which means it controls the gradient of the transfer function.

Training a network based on limited samples is a loosely framed problem. There are many potential models that can be used to train the model such that its output is very close to the expected output. In order to choose one of the possible alternatives, the problem needs to be regularized. That is, additional conditions apart from the requirement that the response of the trained network must agree with the expected one need to be imposed.

Bayesian regularization is one technique that is commonly employed to achieve this objective. In Bayesian regularization, the additional objective imposed ensures that the selected trained network not only minimizes a metric of the error but also achieves this with weights that are of as small a magnitude as possible. In this research, we propose Bayesian regularization instead of back propagation for the fine tuning process after unsupervised training. The objective function employed is as follows,

$$F_W=\alpha P+\beta E_W \qquad (8)$$

where $F_W$ is the new objective function in the process of supervised training, P is the original one as per Equation (9). $E_W$ is the Bayesian regularization item, and α and β are performance parameters that need to be calculated in the iterations or be set before iteration. $E_W$ has the form of mean square of weights.

$$P = \frac{1}{T}\sum_{t=1}^{T}(O_t - O_r)^2 \quad (9)$$

$$E_W = \frac{1}{m \cdot n}\sum_{j=1}^{m}\sum_{i=1}^{n} w_{ij}^2 \quad (10)$$

where, T is the testing set, $O_t$ is the output at testing set t, $O_r$ is the ideal output or teacher's signal, and $w_{ij}$ is the weight between layer i and j. If α≫β, then the first part of $F_W$ dominates, which means that the objective of the training is to decrease the training error. Specifically, if α=1, β=0, then $F_W$=P, and the Bayesian regularization becomes ordinary back propagation. On the other hand, if α≪β, the training will focus on decreasing the weights. So, by introducing this regularization item, one can expect that weights that do not contribute to the response will be minimized ensuring thus that only parts of the network that have learned "important" features common to all the input patterns will remain. Therefore, an improvement in the response of the trained network to unknown test inputs is expected.

Bayesian regularization models are then trained by calculating values of α and β during the training process. During this process, the weights are treated as random variables, and assume that the prior probabilities of P and $E_W$ are Gaussian. α and β can then be obtained by using Bayes criterion.

In the following two case studies, we use Matlab as our computing tool, and the code we are using was obtained from www.deeplearning.net and has been improved for our model and experiments.

## IV. CASE STUDIES

To evaluate the performance of Regularized DBN for road safety applications, two case studies are conducted, focusing on prediction of collisions on individual short sections of Highway 401 of Ontario and a set of maintenance routes in Ontario. Different road conditions and data are collected and processed before being used for model calibration and testing. Additionally, these experiments are also used to explore the relationship between model performance, data size and model structure. Two traditional models - negative binomial (NB) and kernel regression (KR) [6], and one Bayesian neural network - the improved version of the BP neural network [36] are included for performance benchmarking in both case studies.

### A. Case Descriptions

*Case 1, Highway 401 Ontario*

This case study is based on historical collisions and related data from Highway 401 in Ontario, Canada. This highway, which plays a crucial role for the region's socio-economic development, is one of the busiest highways in North America and connects Quebec in the east and the Windsor-Detroit international border in the west. The total length of the highway is 817.9 km of which approximately 800 km is selected for this study. According to 2008's traffic volume data, the annual average daily traffic (AADT) ranges from 14,500 to 442,900 indicating comparatively a very busy road corridor.

The databases used in this case study include: 1) historical crash records for the period from 2000 to 2008 extracted from MTO's Accident Information System (AIS); 2) historical AADT data for the same years from MTO's Traffic Volume Inventory System (TVIS); and 3) road geometric features from MTO's Highway Inventory Management System (HIMS) database. Note that each record in this database is referenced to MTO's linear highway reference system (LHRS). LHRS is a one-dimensional spatial referencing system with a unique five-digit number representing a node/link on a particular highway. LHRS can be used to locate the position of features on a map using a Geographical Information System (GIS) tool.

Traffic count data consists of AADT and average annual commercial vehicle counts for the period 2000 - 2008. As each observation records the LHRS and offset information, traffic counts can be spatially located using a linear referencing GIS tool. The whole highway is divided into 418 homogeneous sections (HS) with length ranging from 0.2 km to 12.7 km. Each HS is then assigned the nearest traffic observation. Ae total of 170 traffic counting stations were available for the 418 HSs. Approximately 85% of the HSs have traffic values assigned from a count station less than 2 km away, indicating that the traffic data is quite an extensive in this particular study area. Finally, the processed crash and traffic data are integrated into a single dataset with HS and year as the mapping fields, resulting in a total of 3762 records. The selected input features used in the study are exposure, AADT, left of shield, median width, right of shield, and curve deflection.

*Case 2, Ontario Highways*

Case study 2 focuses on modelling collisions that occurred under adverse winter weather conditions on a set of highways in Ontario, Canada. The highways are structured into 31 patrol routes, varying from 12.9 to 139.5 kilometers in length, as the basic spatial analysis unit. Hourly collision data, along with traffic count, weather, and road surface conditions data for six winter seasons (2000-2006) are compiled for this study. A brief description on data sources and processing steps is given below. In order to evaluate the performance of deep learning model, the first four years' data are used as the training set (85183), and the last two years' data as the testing set (36875).

Collision data comes from Ministry of Transportation, Ontario (MTO), and is originally collected by the Ontario Provincial Police. This database includes information about each collision at personal level including collision time, collision location, collision type and severity, weather and road surface conditions. Hourly traffic count data is extracted from loop detector data obtained from MTO's COMPASS system and permanent data count stations. The average value is taken for highway sections with multiple count data. Similarly, hourly weather data such as temperature, precipitation, visibility, wind speed are collected from nearby Road Weather Information System and Environment of Canada weather stations. The Road Surface Index (RSI) variable is constructed as a surrogate measure based on MTO's road surface condition weather information system. It measures the frictional levels of road sections. All these data sets are merged into a single hourly data set using date, time and location as the basis of merging for each selected highway section. Finally, only the hours defined by snow storm events for the given six winter seasons are considered.

In these case studies, the performance of each model is estimated based on mean absolute error (MAE) and root mean square error (RMSE), as defined in Equations (11-12).

$$\text{MAE} = \frac{\sum_{i=1}^{n} |P_i - O_i|}{n} \quad (11)$$

$$\text{RMSE} = \sqrt{\frac{\sum_{i=1}^{n} (P_i - O_i)^2}{n}} \quad (12)$$

where, $P_i$ is the predicted collision frequency for $i^{th}$ observation, $O_i$ is the $i^{th}$ observed collision frequency, and n is the total number of observations.

Note that the difference between these two performance measures, MAE and RMSE, is how the residuals are weighted. In MAE, equal weights are given to the residuals from the observed points, whereas in RMSE larger residuals are given greater weights by squaring the deviation. Hence, the smaller the magnitude is, the better the performance level we achieve.

### B. Regularized DBN Model – Model Settings

Just like general DBN model, a Regularized DBN must also be fine-tuned in terms of model structure and learning rate before being able to achieve good performance. First of all, for a given Regularized DBN, the learning rate for its unsupervised component must be specified. This parameter controls how much the weights between layers should be updated in each iteration. Selection of an appropriate learning rate is very important. If the learning rate is too large, the reconstruction error (the index to evaluate unsupervised learning) and weights usually increases dramatically. On the other hand, if learning rate is too small, more epochs are needed, which not only increases the learning time but may also limit its chance to find the global optimal solutions.

Figure 2 shows the training error as a function of the number of training iterations for Case 1, which provides evidence on the minimum number of training iterations that should be used before the learning process reaches a plateau. The structure pertaining to a Regularized DBN model is defined by the number of hidden layers and number of units (or nodes) in each layer, which could have a significant effect on the performance of the model. Too many hidden layers and hidden unites can increase the computation and training time as well as leading to overfitting while too few hidden layers and hidden units may lead to poor feature learning and under-fitting. The significance of these parameters on the model's performance means that these values must be chosen carefully. Previous studies have shown that the most effective method for this is an empirical one.

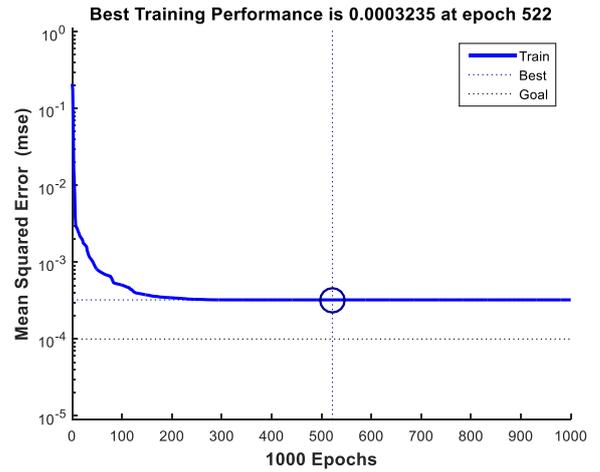

Fig.2. Fine tuning error in Regularized DBN (Case 1)

In addition to the learning rate and model structure, the input features themselves are also important to the performance. Normally, it is best to include as many good features as possible, however, some features may contain too much noise and their inclusion may reduce the accuracy of the model. To assess these effects, this paper also explored the relationship between features and performance. In case 2, in order to test how road features can affect the final modelling result, we designed two experiments. The first one includes the six factors which have been identified in our previous study [6] as the significant factors influencing winter road safety, namely, traffic exposure (combination of traffic volume and length), temperature, precipitation, visibility, RSI and wind speed. The second experiment includes all variables with available data - 16 variables: region, road type, storm hour, monthly ID, temperature, wind speed, visibility, hourly precipitation, RSI, WRM, anti-icing, traffic volume, length, kilometers-fully paved shoulder, kilometers - partially paved shoulder, number of t intersections to be swept, number of bridges.

For Case Study 1, we use a network structure of 6-10-10-1 (6 units for the input layer - 10 units for the hidden layer 1 - 10 units for the hidden layer and 2 - 1 units for the output layer) for the model. This structure is selected on the basis of our previous work [37], which finds that when the training dataset is around 3,000 two hidden layers with no more than 10 are the most effective. Also, different learning rates varying from 0.5 to 3 are tried, with the learning rate of 1.0 being the most effective. In terms of training, we use a total of 20 unsupervised iterations and 1000 fine-tuning iterations (as shown in Figure 2) to ensure model convergence. In Case Study 2, the training dataset is much larger than the one in Case Study 1; as a result, the size of hidden layers should be larger. Model structures with 10 to 40 hidden neurons are tested, of which the structure with 30 hidden neurons is found to outperform the other structures. So the structure of the model is set to be 16-30-30-1. Accordingly, a total of 50 unsupervised iterations and 500 fine-tuning iterations are adopted. A learning rate of 2.0 is selected after trying a range of values from 1 to 5.

## C. Model Performance

### Case 1, Highway 401 Ontario

We first analyze the effect of data size on the performance of the different models. The following bootstrapping process is followed:

1) Split the given dataset into two subsets: a training set and a testing set. The training set includes the first seven years of data (2000-2006) while the testing set includes the remaining two years of data (2007-2008).

2) A subset of data at a specific percentage is randomly drawn from the training data set.

3) The subset of data of a specific size (varying from 5% to 100% of the total training set) is then used to calibrate or train the candidate models, which is subsequently used to predict the collisions at the testing data set. The MAE and RMSE are then calculated.

4) Repeat Step 2) to 3) for 100 times. After done, calculate the average, minimum and maximum MAE and RMSE of the 100 repetitions.

5) Repeat the previous steps for all subset size.

As discussed previously, Bayesian neural network (Bayesian NN), which is the improved version of traditional BPNN using Bayesian function to calculate the output, is also included in the subsequent comparison. Note that the Regularized DBN also employed the Bayesian function in its fine tuning process. In this case, the same training parameters for both models are used (as described in 4.2). The training and testing results are shown from Figure 3 to Figure 5. Figure 3 is the comparison of average MAE and RMSE of different models. When the size of the training data increases, the performance of Bayesian NN and Regularized DBN are both improving. Regularized DBN does not act as well as Bayesian NN at first, but pulls up and finally outperforms Bayesian NN after dataset size reaches 60%. The performance pattern shown in this figure also suggests that Regularized DBN is more stable than Bayesian NN because of the unsupervised training process.

Figure 4 and Figure 5 show a comparison of performance between Regularized DBN and three commonly used models, including NB, KR and Bayesian NN as related to data sizes. The performance of NB does not seem to change substantially as training data increases. Similarly, in both KR and Bayesian NN, the best results show some improvement, but eventually reach a limit. In contrast, the performance of Regularized DBN model continues to improve as data size increases, and when the training data becomes large enough, it significantly outperforms the other models.

Figure 6 is the test MAE and RMSE with the training data increasing from 5% to 100%. Each box has the same meaning as before, the central mark is the median, and the edges of the box are the 25th and 75th percentiles. MAE and RMSE are high at low data sizes, but decrease quickly as the data size increases. It levels off after the data size reaches 50%. This result suggests that the performance of Regularized DBN model improves quickly as data increases. Diminishing returns are visible after the data size reaches 50%, as the MAE and RMSE display lower rates of change. Figure 6 also shows that the performance of Regularized DBN model is sensitive to the amount of training data, and as the training data increases, the performance also improves.

Table 1 shows the values of test MAE (minimum, average, maximum) and RMSE (minimum, average, maximum) of Regularized DBN in Figure 5, while Table 2 is the result of testing MAE and RMSE using different models with whole training data in Figure 4. The final MAE is 8.00 and RMSE is 15.24 using deep learning, which are all better than the other models. The improvements are 32.20% and 42.67% respectively, comparing to NB. It is found that the Regularized DBN model outperformed KR - another non-parameter data-driven model used in our previous study [6], by 9.62% in MAE and 14.60% in RMSE,

TABLE I. PERFORMANCE OF REGULARIZED DBN

| Data | MAE_Min | MAE_Max | MAE_Ave | RMSE_Min | RMSE_Max | RMSE_Ave |
|---|---|---|---|---|---|---|
| 5% | 13.8631 | 32.6482 | 19.2232 | 32.1981 | 87.7819 | 49.2162 |
| 10% | 13.1077 | 28.2934 | 18.6266 | 34.4220 | 79.2412 | 49.8757 |
| 15% | 13.1236 | 28.8359 | 17.9264 | 31.2016 | 80.5257 | 48.2687 |
| 20% | 12.8877 | 27.4936 | 18.4424 | 33.8809 | 77.3719 | 51.6939 |
| 25% | 12.3472 | 25.0407 | 18.4220 | 25.8497 | 71.8663 | 52.2952 |
| 30% | 12.0857 | 25.9500 | 17.4904 | 31.7284 | 78.1760 | 50.5779 |
| 35% | 13.1557 | 23.4138 | 16.2206 | 31.1332 | 67.7740 | 46.9246 |
| 40% | 11.1993 | 20.8038 | 15.0942 | 28.9043 | 63.9712 | 43.6239 |
| 45% | 10.9580 | 19.1842 | 14.2455 | 24.6492 | 61.2754 | 41.1000 |
| 50% | 10.4449 | 18.3780 | 13.7026 | 25.3932 | 54.4649 | 39.1953 |
| 55% | 10.1305 | 18.3707 | 12.9291 | 22.1252 | 55.3869 | 36.3826 |
| 60% | 9.7583 | 24.1627 | 12.6000 | 20.8563 | 56.1061 | 34.9963 |
| 65% | 8.9629 | 16.9705 | 11.9375 | 19.8250 | 53.9345 | 33.2026 |
| 70% | 8.8315 | 15.4428 | 11.7109 | 18.1788 | 48.5473 | 32.1794 |
| 75% | 8.5076 | 15.0573 | 11.2854 | 17.0575 | 46.1240 | 30.4296 |
| 80% | 8.2508 | 13.9346 | 10.9134 | 15.5402 | 44.4648 | 28.4161 |
| 85% | 8.8127 | 13.1249 | 10.6168 | 17.6840 | 43.2967 | 27.7842 |
| 90% | 8.3327 | 13.6888 | 10.5571 | 17.3014 | 42.9242 | 28.0256 |
| 95% | 8.5425 | 13.5272 | 10.6435 | 17.7901 | 42.6633 | 27.5803 |
| **100%** | **7.9982** | **13.5862** | **10.2054** | **15.2446** | **43.5867** | **26.6313** |

TABLE II. COMPARISON OF MODELS

| Models | NB (Base) | KR | | Bayesian NN | | **Regularized DBN** | |
|---|---|---|---|---|---|---|---|
| | | Error | %Improvement | Error | %Improvement | **Error** | **%Improvement** |
| MAE | 11.80 | 8.85 | 25.00 | 8.60 | 27.12 | **8.00** | **32.20** |
| RMSE | 26.60 | 17.85 | 32.89 | 16.51 | 37.93 | **15.24** | **42.67** |

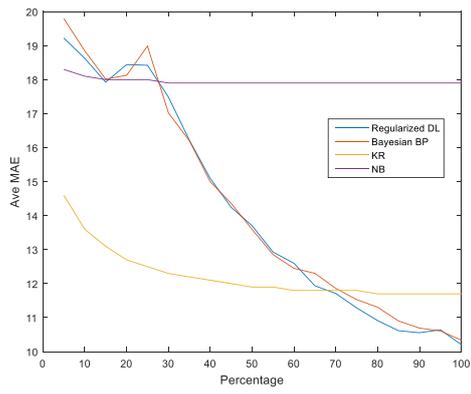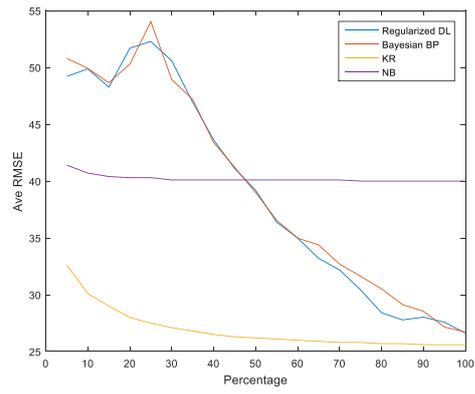

Fig.3. Average performance of different models by training data set sample size (% of total training data)

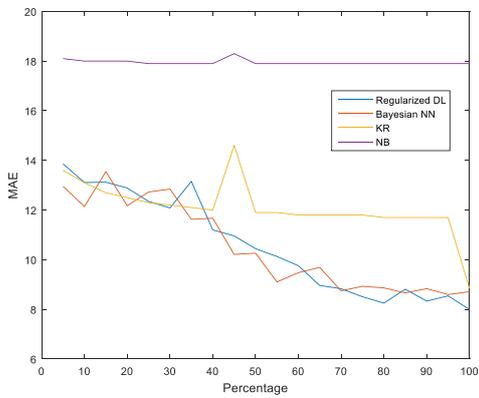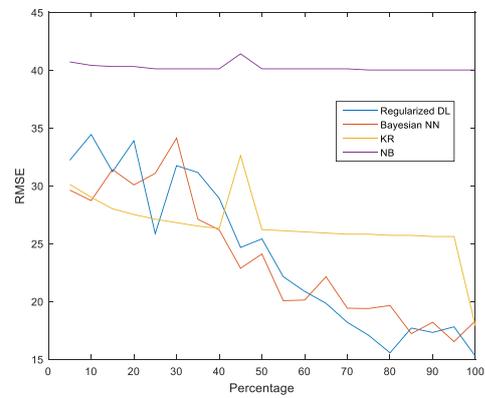

Fig.4. Best performance of different models by training data set sample size (% of total training data)

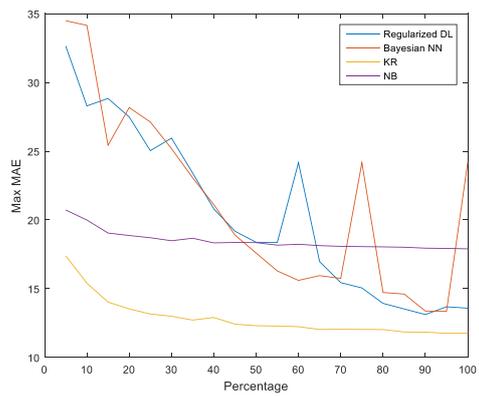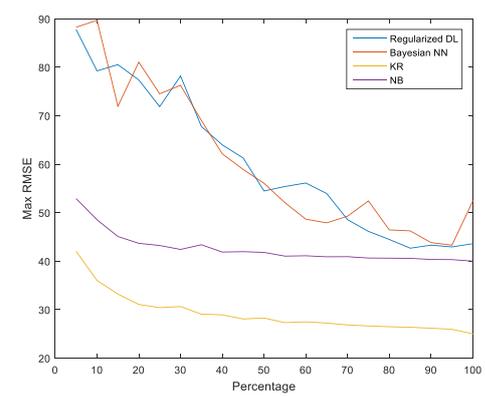

Fig.5. Worst performance of different models by training data set sample size (% of total training data)

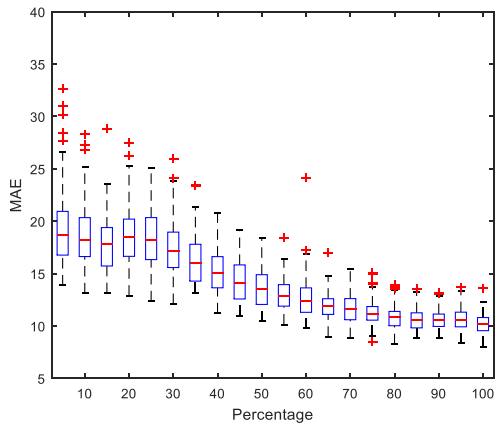 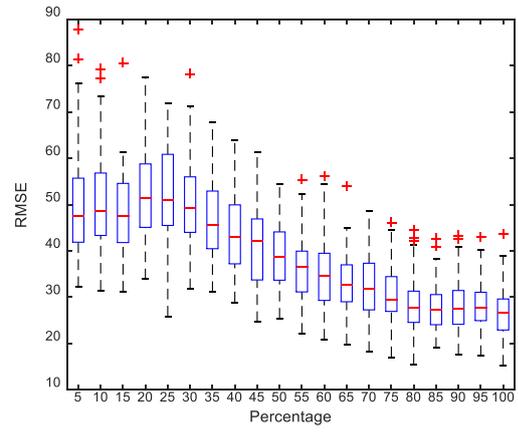

(a)

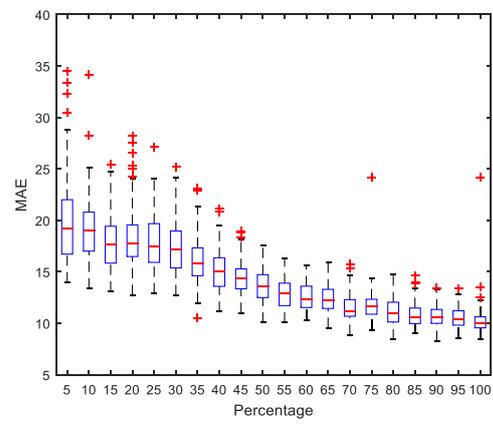 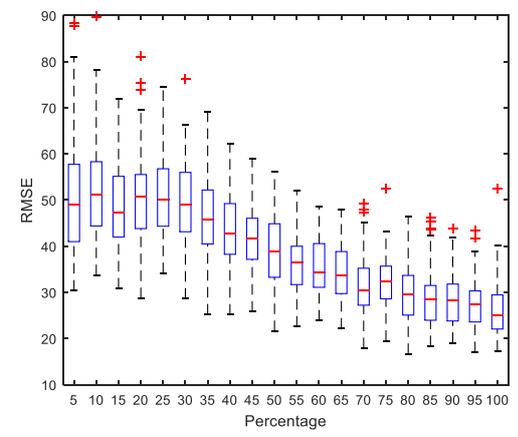

(b)

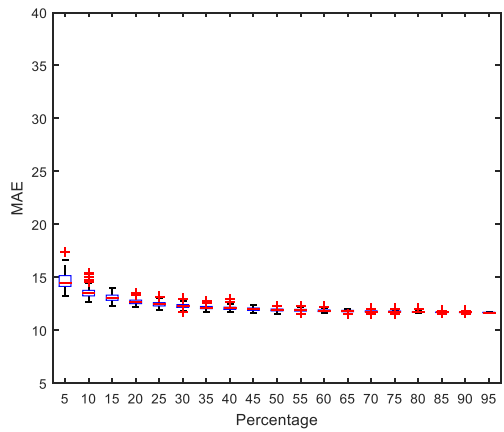 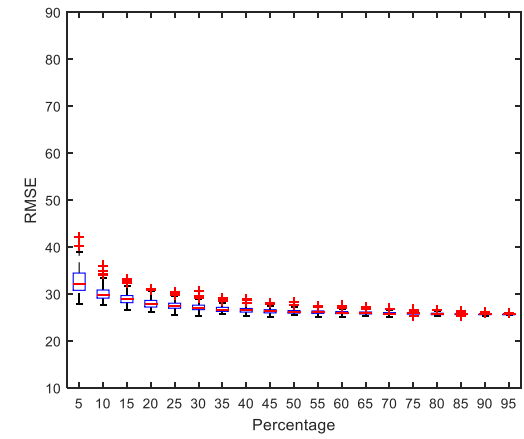

(c)

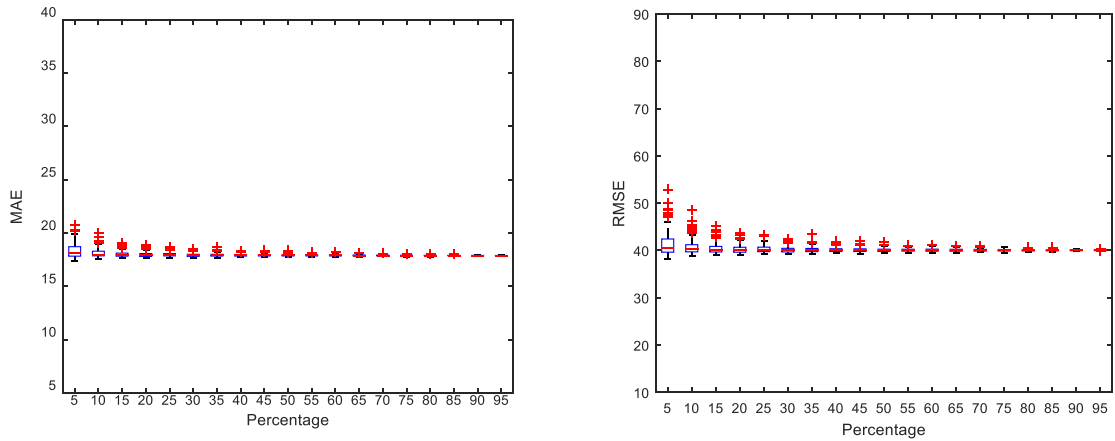

(d)

Fig.6. MAE and RMSE by training data set sample size using different models (% of the total training data set, a- Regularized DBN, b- Bayesian NN, c-KR, d-NB)

*Case 2, Ontario Highways*

We implement Regularized DBN with two hidden layers in this case. Different network structures with varying number of nodes in each of the hidden layers are evaluated to find out the relationship between performance, network size and training data. As the dataset is very large - 122,058 observations in total in this case, we start with taking a random subset of 1% of the data for training, and increase it by 1% each time until 100% of the data has been used. For each data size, the process is repeated 25 times, with each taking a random subset, to reduce the effect of the randomness in the training data on model performance. The training and testing results of using 30 hidden neurons are shown from Figure 7 to Figure 9.

Figure 7 shows the boxplot of the MAE of the model as a function of data size. The results have clearly shown that the amount of training data has an important effect on the performance of the model. The MAE is initially very high at low data size, with relatively large error margin (indicated by the blue rectangle). After the data size reaches 30% of the training dataset, the MAE drops dramatically and becomes very stable. This is likely caused by the deep learning algorithm reaching a limit and learning most of the useful knowledge after using 30% of the data; and consequently, the test error reaches the plateau.

Figure 9 is the average and variance of MAE and RMSE of the estimates from Regularized DBN. When the data size is small, it may not encapsulate the major features affecting collision occurrences, which makes the testing result unstable, as shown by the ups and downs in MAE. However, when the selected training set is large enough (over 30%), the testing error becomes much stable, especially the variance of MAE becomes extremely small. Similar patterns can be observed with respect to RMSE (Figure 8 and Figure 9).

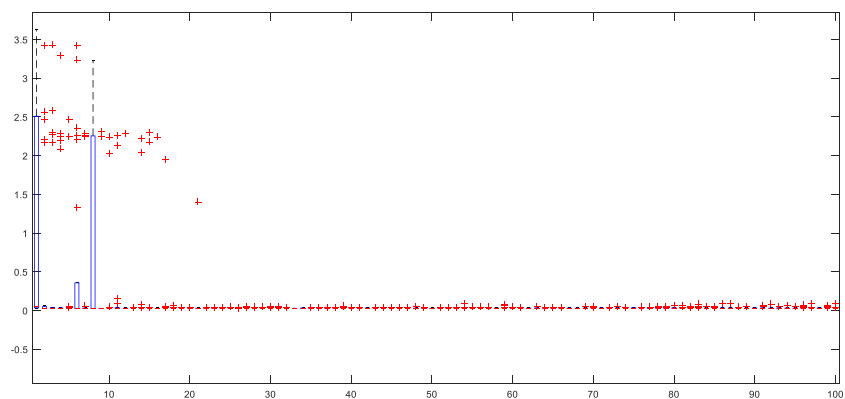

Fig.7. MAE as a function of data size (% of the total training data set)

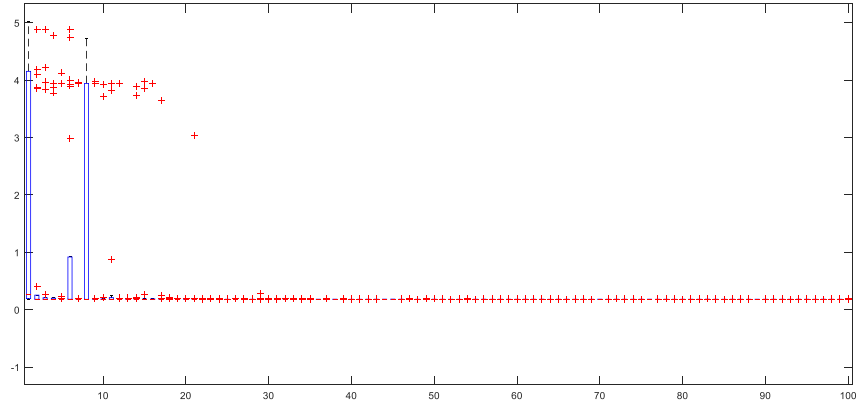

Fig.8. RMSE as a function of data size (% of the total training data set)

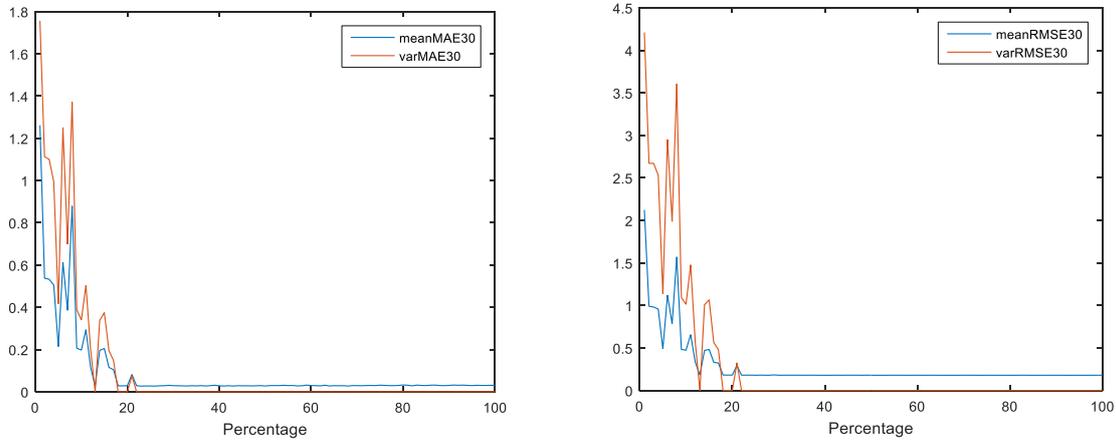

Fig.9. Average and variance of MAE and RMSE by Regularized DBN

TABLE III. COMPARISON OF DIFFERENT MODELS AND PARAMETERS

| Model | NB | KR | Regularized DBN_ 6 features | Regularized DBN_ 16 features |
|---|---|---|---|---|
| MAE | 0.05 | 0.047 | 0.0238 | **0.0238** |
| RMSE | 0.187 | 0.1770 | 0.1772 | **0.1760** |
| MAE _Ave | / | / | 0.0273 | **0.0304** |
| RMSE _Ave | / | / | 0.1802 | **0.1797** |

The results of using different models and parameters are summarized in Table 3. We can conclude that a Regularized DBN with 30 hidden neurons has the best prediction power, having the lowest MAE of 0.0238, compared to 0.05 by NB and 0.047 by KR. Similarly, the best RMSE is 0.1760, compared to 0.187 by NB and 0.1770 by KR. When using 100% training dataset, once NB and KR are trained, the results will not be changed, so there is no average MAE or average RMSE for them in Table 3. When using 16 features, MAE is a little worse than when using only six features. This is because more features mean more uncertainty and sometimes unwanted features. However, in terms of RMSE, the model with 16 features is better than the one with six features, indicating that a model with more features has more predictive power for cases with extreme values. This result is a consequence of the fact that larger model sizes need more training time and data and may learn less important features.

V. CONCLUSIONS

In this paper, we proposed a machine learning approach for modelling road collisions. The research has made two main contributions. First, a Regularized DBN model is introduced as an alternative to traditional parametric and nonparametric models for predicting expected collision risk of highways. This model is able to receive and process continuous real world data by using Gaussian input units. The fine-tuning method of Bayesian regularization is employed to reduce the over fitting problem common to machine learning models. Additionally, model training parameters such as network size and data quantities used are also analyzed in terms of their impact on the training process and the final model accuracy.

Secondly, the research has conducted an extensive empirical investigation of the performance of Regularized DBN as compared to the traditional models. Two case studies have been conducted and the results have shown that the proposed model is capable of predicting collision frequencies with much higher accuracy than the models that have commonly been used in the literature. Furthermore, it requires much less time to train and have more flexibility to make use of available data for improved model accuracy.

This research represents an initiating effort with several unsolved questions that need to be investigated in the future. For example, the robustness of the model performance to network structure and size needs to be further studied so that methods for determining the best network configuration could be devised for specific problems. The second issue is related to the need of adapting network structure to increased data size. Our preliminary analysis has shown the performance advantage of a fixed DBN model as data size increases. There may be further performance gain with adjusted model structure. Lastly, the general modelling approach we have adopted follows the traditional method of aggregating collision data over specific spatial units (e.g., intersections or uniform segments) and time interval (e.g., yearly), which is expected to lead to significant loss of information. There is a need to develop a different modelling paradigm that can take the full advantage of the learning power of the deep learning models.

ACKNOWLEDGMENT

This research is jointly supported by Ontario Ministry of Transportation (MTO) and National Science and Engineering Research Council of Canada (NSERC).